\RequirePackage{fix-cm}
\documentclass[natbib,twocolumn]{svjour3}     
\smartqed  
\usepackage{graphicx}
\usepackage{epstopdf}
\usepackage{amssymb}
\usepackage{rotating}
\usepackage{paralist}
\usepackage{bm}
\usepackage{booktabs}
\usepackage{relsize}
\usepackage[utf8]{inputenc} 
\usepackage{xspace}
\usepackage[linesnumbered,ruled,vlined]{algorithm2e}
\usepackage{url}
\usepackage{amstext}
\usepackage{amsmath}
\usepackage{enumitem}
\usepackage[final]{changes}
\setcitestyle{aysep={}}
\usepackage{breakcites}

\newcommand{\Qone}{\ensuremath{\text{Q}_1}\xspace}
\newcommand{\Qtwo}{\ensuremath{{\text{Q}_2}}\xspace}

\newcommand{\ASone}{\ensuremath{{\text{A}_1}}\xspace}
\newcommand{\AStwo}{\ensuremath{{\text{A}_2}}\xspace}
\newcommand{\fdist}{\ensuremath{{d}}\xspace}

\newcommand{\vece}{\ensuremath{\mathbf{e}}\xspace}

\newcommand{\peig}{\ensuremath{p}\xspace}
\newcommand{\neig}{\ensuremath{n_{\lambda+}}\xspace}
 \journalname{Soft Computing}
\begin{document}

\title{An Empirical Approach For Probing the Definiteness of Kernels*\thanks{*preprint submitted to Soft Computing journal, Springer}
}


\author{\mbox{Martin Zaefferer} \and  \nolinebreak
\mbox{Thomas Bartz-Beielstein} \and   \nolinebreak
\mbox{G{\"u}nter Rudolph}
}

\authorrunning{ M.\ Zaefferer, T.\ Bartz-Beielstein and G.\ Rudolph} 

\institute{Martin Zaefferer (corresponding author) \at
	Faculty of Computer Science and Engineering Science,\\ TH Köln - University of Applied Sciences, \\
	Steinm{\"u}llerallee 1, 51643 Gummersbach, Germany\\
	Telephone: +49 2261 8196 6327  \\
	\email{Martin.Zaefferer@th-koeln.de} 
	\and
	Thomas Bartz-Beielstein \at		 
	Faculty of Computer Science and Engineering Science,\\ TH Köln - University of Applied Sciences,\\
	Steinm{\"u}llerallee 1, 51643 Gummersbach, Germany\\
	\and
  G{\"u}nter Rudolph \at
	Department of Computer Science,	
	TU Dortmund University \\
	Otto-Hahn-Str. 14, 44227 Dortmund, Germany\\
}

\date{Received: date / Accepted: date}

\maketitle
\begin{abstract}
Models like support vector machines or 
Gaussian process regression \added{often} require positive 
semi-definite kernels. \added{These kernels} may be based 
on distance functions. 
While definiteness is 
proven for common distances and kernels, a proof for a new kernel may 
require too much time and effort for users who simply aim at practical usage.
Furthermore, designing definite distances or kernels may be equally 
intricate.
Finally, models can be enabled to use indefinite kernels. 
This may deteriorate the accuracy or computational cost of the model. 
Hence, an efficient method to determine definiteness is required.
We propose an empirical approach. We show that sampling as well as 
optimization with an evolutionary algorithm may be employed
to determine definiteness.
We provide a proof-of-concept with 16 different distance measures for permutations.
\added{Our approach allows to disprove definiteness if a respective counter-example is found.
It can also provide an estimate of
how likely it is to obtain indefinite 
kernel matrices}.
This provides 
a simple, efficient tool to decide
whether additional effort should be spent on 
designing/selecting a more suitable kernel or algorithm. 

\end{abstract}

\section{Introduction}\label{sec:intro}


The definiteness of kernels and distances is an important issue
in statistics and machine learning~\citep{Feller1971,Vapnik1998,Schoelkopf2001}.
One application that recently gained interest is the field of surrogate model-based
combinatorial optimization~\citep{Moraglio2011,Zaefferer2014b,Bartz-Beielstein2016n}.
Continuous distance measures are replaced by distance measures that are adequate 
for the respective search space (e.g., permutation distances or string distances).
Such a measure will have an effect on the definiteness of the employed kernel function. 
For arbitrary problems, practitioners may come up with any kind of suitable distance measure or kernel.

While it is easy
to determine definiteness of matrices, determining the definiteness of a function is not as simple. 
Proving definiteness by theoretical means may be infeasible in practice
(\citeauthor{Murphy2012} \citeyear{Murphy2012}, p.~482; \citeauthor{Ong2004} \citeyear{Ong2004}).
It may be equally difficult to design a function to be definite.
Finally, algorithms may be adapted to handle indefinite kernels.
These adaptations usually have a detrimental impact on the computational effort 
or accuracy of the derived model.
Hence, this study tries to answer 
the following two research questions:
\begin{description}
\item[\Qone] \textbf{Discovery:} Is there an efficient, empirical approach to determine the 
definiteness of kernel functions based on arbitrary distance measures? 
\item[\Qtwo] \textbf{Measurability:}  If \Qone can be answered affirmatively, 
\added{can we quantify to what extent a lack of definiteness is problematic?}
\end{description}
\Qone tries to find an answer to the general question of whether or not a function is definite.
\added{Measurability} (\Qtwo) is important, as 
this may allow \added{determining} the impact that indefiniteness has in practice. 
For instance, if a kernel rarely produces indefinite matrices, an optimization or learning process that explores
only a small subset of the search space may not be negatively affected.
Therefore, this article proposes two approaches.

\begin{description}
\item[\ASone]  \textbf{Sampling:}  Random sampling is used to determine the proportion of solution sets with
indefinite distance or kernel matrices for a given setting. 
\item[\AStwo] \textbf{Optimization:} Maximizing the largest eigenvalue related to a certain solution set with an Evolutionary Algorithm (EA), hence finding
indefinite cases even when they are rare. 
\end{description}

If \added{either approach detects an indefinite matrix}, the respective kernel function is demonstrated to be
indefinite. On the other hand, if no indefinite matrix is detected,
definiteness of the function is not proven. Still, this indeterminate result
may indicate that indefiniteness is at least unlikely to occur. Hence, the function may be treated as unproblematic 
in practical use.

Section~\ref{sec:back} provides the background of the methods presented in
Section~\ref{sec:meth}.
The experimental setup for a proof-of-concept, based on permutation distance measures, is described in Section~\ref{sec:exp}.
Results of the experiments are analyzed and discussed in Section~\ref{sec:results}.
Finally, a summary of this work as well as an outlook on future research directions is given in Section~\ref{sec:summary}.

\section{Background: Distances, Kernels and Definiteness}\label{sec:back}

\subsection{Distance Measures}\label{sec:dist}
Distance measures compute the dissimilarity of two objects  $x,x' \in \mathcal{X}$,
 where we do not assume anything about the nonempty set $\mathcal{X}$.
Such objects can \added{be, e.g.}, permutations, trees, strings, or vectors of real values.
Thus, a distance measure 
$\fdist: ~ \mathcal{X} \times \mathcal{X} \to \mathbb{R}_+$ 
expresses a scalar, numerical value $\fdist(x,x')$ that should become
larger the more distinct the objects $x$ and $x'$ are.  
For a set of $n \in \mathbb{N}$ objects, the distance matrix $D$ of dimension $n \times n$
collects all pairwise distances $D_{ij}=\fdist(x_i,x_j)$ with $i=1, \ldots ,n$ and $j=1, \ldots ,n$.
Intuitively, distances can be expected to satisfy certain conditions, e.g., they 
should be zero when comparing identical objects.

A more formal definition is implied by the term \textit{distance metric}.
A distance metric $\fdist(x,x')$ is symmetric $\fdist(x,x')=\fdist(x',x)$, non-negative $\fdist(x,x')\geq0,$
preserves identity $\fdist(x,x')=0 \iff x=x' , $ and satisfies the triangle inequality 
$\fdist(x,x'') \leq \fdist(x,x') + \fdist(x',x'').$
Distance measures that do not preserve identity are often called pseudo-metrics.

An important class of distance measures are edit distance measures.
Edit distance measures can be defined to count the minimal number of edit operations 
required to transform one object  into another. An edit distance
measure may concern one specific edit operation (e.g., only swaps)
or a set of different operations (e.g., Levenshtein distance with substitutions, deletions, insertions).
Edit distances usually
satisfy the metric axioms.


\subsection{Kernels}
In the following, a kernel (also: kernel function, similarity measure or correlation function) is defined as 
a real valued function $k(x,x')$ with 
\begin{equation}\label{eq:kernel}
\begin{split}
k: &~ \mathcal{X} \times \mathcal{X} \to \mathbb{R}\\
 &~(x,x') \mapsto k(x,x')
\end{split}
\end{equation}
that will usually be symmetric $k(x,x')=k(x',x)$ and
non-negative $k(x,x')\geq 0$~\citep[p. 479]{Murphy2012}.
Kernels can be based on distance measures, i.e., $k(\fdist(x,x'))$.

\subsection{Definiteness of Matrices}\label{sec:def} 
One important property of kernels and distance measures is their definiteness.
We refer to the literature for more in-detail descriptions and proofs, which are
the basis for the following paragraphs~\citep{Berg1984,Schoelkopf2001,Camastra2007}. 

First, we introduce the concept of matrix definiteness.
A symmetric, square matrix $A$ of dimension $n\times n$ ($n \in \mathbb{N}$) is positive definite (PD) if and only if
\begin{equation*}\label{eq:pdmat}
\sum_{i=1}^n \sum_{j=1}^n c_i c_j A_{ij} > 0,
\end{equation*}
for all $c \in \mathbb{R}^n\setminus\{0\}$. This is equivalent to 
all eigenvalues $\lambda_1 \leq \lambda_2 \leq \ldots \leq \lambda_n  $  of the matrix $A$ being positive.
Due to symmetry, the eigenvalues are $\lambda \in \mathbb{R}^n$.
Respectively, a matrix is negative
definite (ND) if all eigenvalues are negative. 
If eigenvalues are non-negative (i.e., some are zero) or non-positive, the matrix is respectively
called Positive or Negative Semi-Definite (PSD, NSD). 
If mixed signs are present in the eigenvalues, the matrix may be called indefinite.
Kernel (or correlation, covariance) matrices are examples of matrices that
have to be PSD.

A broader class of matrices are Conditionally PSD or NSD (CPSD, CNSD).
Here, the coefficients satisfy  
\begin{equation}\label{eq:conditional}
\sum_{i=1}^n c_i = 0,
\end{equation}
with $n>1$.
All PSD (NSD) matrices are CPSD (CNSD).
To check conditional definiteness, let  the $n \times n $ matrix $P$ be 
\begin{equation*}\label{eq:augmentedD3}
P=
 \begin{pmatrix}
I_{(n-1)} - \vece \vece^T /n ~~~~&  \vece / n \\ 
(0, \ldots, 0)~~~~  & 1
 \end{pmatrix}
\end{equation*}
with $\vece=(1, \ldots, 1)^T$, and 
$ 
B= PAP^T
$. 
Then, $A$ is CNSD if and only if 
\begin{equation}\label{eq:augmentedD2}
\hat{A}= B_{n-1}
\end{equation}
is NSD~\citep[Algorithm 1]{Ikramov2000}. 
Here, $B_{n-1}$ is the leading principal submatrix of $B$, that
is, the last column and row of $B$ are removed.

\subsection{Definiteness of Kernel Functions}\label{sec:kerndef}
In a similar way to the definiteness of matrices, definiteness can also be determined for kernel functions. 
The upcoming description roughly follows the definitions and notations by~\citet{Berg1984}
and~\citet{Schoelkopf2001}.
For the nonempty set $\mathcal{X}$, a symmetric kernel $k$ is called PSD if and only if 
\begin{equation*}\label{eq:PSD}
\sum_{i=1}^n \sum_{j=1}^n c_i c_j k(x_i,x_j) \geq  0,
\end{equation*}
for all $n \in \mathbb{N}$, $x \in \mathcal{X}$ and $\mathbf{c} \in \mathbb{R}^n$.
A PSD kernel will always yield PSD kernel matrices.


An important special case are conditionally definite functions.
Analogous to the matrix case, they observe the respective condition in equation~\eqref{eq:conditional}.
One example of a CNSD function is the Euclidean distance.
The importance of CNSD functions is also due to the fact that the distance measure
 $\fdist(x,x')$ is CNSD
if and only if the kernel $k(x,x')=\exp(-\theta \fdist(x,x'))$ 
 is PSD $\forall~\theta > 0$ ~\citep[Proposition 2.28]{Schoelkopf2001}.

It is often stated that kernels must be PSD \citep{Rasmussen2006,Curriero2006}.
Still, even an indefinite kernel function may yield a PSD kernel matrix.
This depends on the specific data set used to train the model~\citep{Burges1998};~\citep[Theorem 2]{Li2004}
as well as the parameters of the kernel function.
Some frequently used kernels are known to be indefinite.
Examples are the sigmoid kernel~\citep{Smola2000,Camps-Valls2004} 
or time-warp kernels for time series~\citep{Marteau2014}.

To handle the issue of a kernel's definiteness, different (not mutually exclusive) approaches can be found in the literature.
\begin{itemize}
\item \textbf{Proving:}
Definiteness of a specific function can be \textit{proven} (or disproven) by theoretical considerations
in some cases (cf.~\citep{Berg1984}). For complex cases, or practitioners this may be an
infeasible approach~\citep{Ong2004}. 
\item \textbf{Designing:}
Functions can be \textit{designed}
to be definite~\citep{Haussler1999,Gaertner2003a,Marteau2014}. 
Especially noteworthy are the so called convolution kernels~\citep{Haussler1999}, as
they provide a method to construct PSD kernels for structured data.
For a similar purpose, \citet{Gaertner2004} show how to design a
syntax based PSD kernel for structured data.
However, convolution kernels may be hard to design~\citep{Gaertner2004}.
Also, kernels and distance measures may be predetermined for a certain application.
\item \textbf{Adapting:}
Algorithms or kernel functions may be \textit{adapted}
to \added{be usable} despite a lack of definiteness.
This, however, may affect the computational effort or accuracy of the derived model.
Some approaches alter matrices (or rather, their eigenspectrum), 
hence enforcing PSDness~\citep{Wu2005,Chen2009,Zaefferer2016b}. 
For the case of SVMs, \citet{Loosli2015} provide a nice comparison of
various approaches of this type and propose an interesting new solution to the issue
of indefinite kernels based on learning in Krein spaces.
A recent survey is given by~\citet{Schleif2015}.
\end{itemize}

Between these three approaches, there is a lack of straightforward empirical procedures,
\added{without resorting to complex theoretical reasoning.
How can the definiteness of a function be determined? And what impact does
a lack of definiteness have on a model?
}

Hence, we propose the two related empirical approaches \ASone and \AStwo introduced in
Sec.~\ref{sec:intro} to fill the gap.
An empirical approach may help to overcome the difficulty of theoretical considerations 
or designed kernels. 
Empirical results may also be a starting point for a more formal approach.
Furthermore, it may give a quick answer on whether or not the algorithm will have to be adapted
for non-PSD matrices
(which, if applied by default, would require additional computational effort and may limit the accuracy of derived models).

\section{Methods for Estimating Definiteness}\label{sec:meth}

We propose an experimental approach
to determine and analyze definiteness (\ASone, \AStwo).
As a test case, we determine the definiteness of a distance-based exponential kernel, $k(x,x')=exp(-\theta d(x,x'))$. 
The kernel is definite if the underlying distance function is CNSD.
For a given distance matrix, CNSDness is determined by the largest eigenvalue \added{$\lambda_{n}$} of $\hat{D}$, 
based on equation~\eqref{eq:augmentedD2}.
$D$ is not CNSD if $\lambda_{n}>0$.
We could also probe the kernel matrix $K$, but in this case the kernel parameter  $\theta$
would have to be dealt with.

Of course, we cannot simply check definiteness of a single matrix $D$,
since this would be only one possible outcome of the respective kernel or distance function.
Hence, a large number of solution sets with respective distance matrices has to be generated 
to determine whether any 
\added{of the matrices are CNSD (research question \Qone: discovery)  
and to what extent this may affect a model
(\Qtwo: measurability).}
For smaller, finite spaces a brute force approach may
be viable.  All potential matrices $D$ can be enumerated and checked for CNSDness.
Since this quickly becomes computational infeasible, we propose
to use sampling or optimization instead.



\subsection{Estimating Definiteness with Random Sampling}\label{sec:sample}
To estimate the definiteness of a distance or kernel function we propose a simple random sampling approach (\ASone).
This approach randomly generates $t\in \mathbb{N}$ sets $X_1, \ldots , X_t$.  
Each set has size $n$, that is, it contains $n \in \mathbb{N}$ candidate samples $X=\{ x_1, \ldots, x_n \}$. 

For each set, the distance matrix $D$ is computed, containing distances between all candidates in the set.
Based on this, $\hat{D}$ is derived from equation~\eqref{eq:augmentedD2}.
Then, the largest eigenvalue $\lambda_{n}$ of $\hat{D}$ is computed.
This eigenvalue determines whether $\hat{D}$ is NSD, and hence whether $D$ is CNSD and $K$ PSD.
This is repeated for all $t$ sets. 
The number of times that the largest eigenvalue is positive ($\lambda_{n}>0$) is retained as 
$\neig$. 
Accordingly, the proportion of non-CNSD matrices is determined with
$\peig = \frac{\neig}{t}.$

Obviously, all distance measures \added{that} yield $\peig > 0$ are proven to be non-CNSD. 
Hence, an exponential kernel based on these measures is also proven to be indefinite.
If $\peig = 0$, CNSDness is not proven or disproven.

In general, the proposed method can be categorized as a randomized algorithm of the complexity class RP \cite[p.~21f.]{MR95}. 
That is, it stops after polynomially many steps, and if the output is ``no'' then the distance measure is non-CNSD
with probability 1, and if the output is ``yes'' then the distance measure is CNSD with some probability 
strictly bounded from zero.   

The parameter $\peig$ is an estimator of how likely a non-CNSD matrix is to occur, for the specified set size $n$.
To determine definiteness, the calculation of $\lambda_n$ \added{of $\hat{D}$} is not mandatory, but it may be useful
to see how close to zero $\lambda_n$ is, to distinguish between a pathological case and 
cases where the matrix is just barely non-CNSD. 
\added{We will show in Section~\ref{sec:modelQuality} that $\lambda_n$ \added{of $\hat{D}$} can be linked to model quality.}

Note, that inaccuracies of the numerical algorithm
used to compute the eigenvalues might lead to an erroneous sign of the largest eigenvalue.
To deal with that, one could try to use exact or symbolic methods or else use
a tolerance when checking \added{whether} the largest eigenvalue is larger than zero. In the
latter case, a matrix $\hat{D}$ is assumed to be non-NSD if $\lambda_n>\epsilon$, where $\epsilon$
is a small positive number.

%
%

\subsection{Estimating Definiteness with Directed Search}
If very few sets $X$ yield indefinite matrices, \ASone may fail
to find indefinite matrices by pure chance. 
In such cases, it may be more efficient to replace random sampling with a directed search (\AStwo).
In detail, a set $X$ can itself be viewed as a candidate solution of an optimization problem. 
The largest eigenvalue $\lambda_{n}$ of the transformed distance matrix $\hat{D}$ is the objective to be maximized.
By maximizing the largest eigenvalue, a positive $\lambda_{n}$ may be found 
more quickly (and more reliably).

This optimization problem is strongly dependent on
the kind of solution representation used. Evolutionary Algorithms (EAs)
are a good choice to solve this problem, because they are applicable
to a wide range of solution representations (e.g., real values, strings, permutations, graphs, and mixed \added{search} spaces).
EAs use principles derived from natural evolution for the purpose of optimization. 
That is, EAs are optimization algorithms based on a cyclic repetition of
parent selection, recombination, mutation, and fitness selection (see e.g.,~\citet{Eiben2003}).
These operations can be adapt\-ed to a large variety of representations, mainly depending
on suitable mutation and recombination operators. The EA in this study will operate as follows:

\begin{itemize}
\item \textbf{Individual: $X$.} A set $X=\{ x_1, \ldots, x_n \}$  with set size $n$ is considered as an individual. 
Set elements $x$ are samples in the actual search or input space, as used throughout Sec.~\ref{sec:back}.
\item \textbf{Search space: $S^n$.} All possible sets $X$ of size $n$, i.e., $X \in S^n$. 
\item \textbf{Population: $Z$.} A population of size $r$, containing $X_k \in Z$ with  $k \in \{1,\ldots,r\}$ and $Z\subseteq S^n$.
\item  \textbf{Objective Function:} 
\begin{align}
\begin{split}\label{eq:objective}
f:~ &S^n  \to \mathbb{R} \\
& X \mapsto \lambda_n
\end{split}
\end{align}
where  $\lambda_n$ is the largest eigenvalue of the transformed distance matrix $\hat{D}$ based on equation~\eqref{eq:augmentedD2}. The objective function $f$ is maximized.
\item \textbf{Mutation: Alteration of an individual.}\\
$X_{\text{new}}=mutation(X)=$\\ $\{x_1,\ldots,x_{j-1},submutation(x_j),x_{j+1},\ldots,x_n, \}$,
with $j\in  \{1,\ldots,n\}$.
For the submutation function, any edit operation that works for a sample $x$ can be chosen:  $x_{new}=submutation(x)=edit(x)$.\\
 For example, in case of permutations, one permutation $x_j \in X$ is chosen and mutated with typical permutation edit-operations (swap, interchange, reversal). 
The specific edit-operation is called submutation operator, to distinguish between mutation of the individual set $X$
and the submutation of a single permutation $x_j \in X$. 
\item \textbf{Recombination: Combining two sets.} For recombination, two sets are randomly split and the parts of both sets are joined to form a new set
of the same size. 
\item \textbf{Repair: Duplicate removal.}
Mutation and recombination may create
duplicates ($x_i=x_j$ with $i\neq j$). In practice, duplicates are not desirable and are irrelevant to
the question of definiteness.
Hence, duplicates are replaced by randomly generated, unique samples $x^* \notin X$.
\item \textbf{Stopping criterion: Indefiniteness proven or budget exhausted.}
The optimization can stop when some solution set $X$ is found which yields $\lambda_n > \epsilon$, where $\epsilon$
is a small positive number. Alternatively, the process stops if a budget of objective function evaluations is exhausted.
\end{itemize}

\section{Experimental Validation}\label{sec:exp}
The proposed approaches can be useful in any case where definiteness of kernels is of interest.
The experiments provide a proof of concept of the proposed approaches.
Hence, we chose to pick a recent application as a motivation for our experiments:
 surrogate-model based optimization in permutation spaces \citep{Moraglio2011,Zaefferer2014c}.

In many real-world optimization problems, objective function
 evaluations are expensive.
Sequential modeling and optimization techniques are state-of-the-art in these settings~\citep{Bartz-Beielstein2016n}.
Typically, an initial model is built at the first stage of the process. The model will be subsequently refined by adding further data
until the budget is exhausted. At each stage of this sequential process, available information from the model is used to determine 
promising new candidate solutions.
This motivates the rather small data set sizes used throughout this study.

\subsection{Test Case: Permutations}\label{sec:perm}

As a proof-of-concept, we selected 16  distance measures
for permutations; see Table~\ref{tbl:measures} for a complete list. 
The implementation of these distance measures is taken from the R-package \texttt{CEGO}\footnote{
The package \texttt{CEGO} is available on CRAN at \texttt{http://cran.r-project.org/package=CEGO}.
}.

Similarly to \cite{Schiavinotto2007} we define
$\Pi^m$ as the set of all permutations of the numbers $\{1,2,\ldots,m\}$.
 A permutation has exactly $m$ elements.
We denote a single permutation with $\pi \in \Pi^m$ and $\pi = \{\pi_1, \pi_2, \ldots, \pi_m \}$ where $\pi_i$ is a
specific element of the permutation at position $i$.  
For example, a permutation in this notation is $\pi =\{3,2, 1, 4,5 \}\in \Pi^5$.
Explanations and formulas (where applicable) for the distance measures are given in appendix A.

\begin{table}[ht]
\centering
\caption{Distance measures for permutations.
Second column lists runtime complexity where $m$ is the number of elements of the permutation. Metric
refers to permutation space; these measures may be non-metric in other spaces. LC is short for Longest Common.}
\label{tbl:measures}
\begin{tabular}{llcc}
 \toprule
name& complexity & metric & abbreviation \\ 
 \midrule
Levenshtein  &   $O(m^2)$  & yes & Lev \\ 
Swap &   $O(m^2)$  &  yes& Swa \\ 
Interchange &   $O(m^2) $ &  yes& Int \\ 
Insert &  $O(m log(m))$  & yes & Ins\\ 
LC Substring &  $O(m^2)$  & yes& LCStr\\
R &   $O(m^2)$ &  yes& R\\ 
Adjacency &  $O(m^2)$  & pseudo& Adj\\ 
Position &    $O(m^2)$ & yes& Pos\\ 
Position$^2$ &    $O(m^2)$  & no& Posq \\ 
Hamming &  $O(m)$ &  yes& Ham\\ 
Euclidean & $O(m)$ &  yes& Euc\\ 
Manhattan & $O(m)$ &  yes& Man\\ 
Chebyshev & $O(m)$ &  yes & Che\\
Lee & $O(m)$ &  yes &Lee\\ 
Cosine & $O(m)$ & no&Cos\\ 
Lexicographic & $O(m log(m))$ & yes& Lex\\  
  \bottomrule
\end{tabular}
\end{table}

\subsection{Random Sampling} 
In a single experiment, 
 $t = 10,000$  sets of permutations are randomly generated.
Each set contains $n$ permutations and each permutation has $m$ elements.
For each set, the largest eigenvalue $\lambda_n$ of $\hat{D}$ is computed based on equation~\eqref{eq:augmentedD2}.
To summarize all $t$ sets, the largest  $\lambda_n$ as well as the ratio $\peig = \frac{\neig}{t}$ are recorded.
The tolerance value used to check whether the largest eigenvalue is positive is $\epsilon$=1e-10.
This process is repeated 10 times, to achieve a reliable estimate of the recorded values.

Two batches of experiments are performed. In the first, all 16 distance measures are
\added{examined}, with $n= \{4 , \ldots , 20\} $ and  $m= \{4 , \ldots , 15\} $. 
In the second batch, larger sizes $n=\{21 , \ldots , 40,45,50,60,70,80,90,100 \} $ are tested,
but the permutations are restricted to $m= \{5 , \ldots , 15\} $ and the distance \added{measures} are only LCStr, Insert, Chebyshev, Levenshtein and Interchange.

\subsection{Directed Search}\label{sec:expdirect}
To be comparable to the random sampling approach, the budget for each EA run is
$10,000$ fitness function evaluations. A run will stop if the budget is
exhausted or if $\lambda_n > \epsilon=10^{-10}$.
The population size of the EA is set to 100. The recombination rate is 0.5, the mutation
rate is $1/m$ and truncation selection is used. 

To identify bias introduced
by the choice of \added{the} submutation operator (which may have a strong interaction with
the respective distance measures), each EA run is performed repeatedly with
three different submutation operators:

\begin{compactitem}
\item \textbf{Swap mutation}:  Transposing two adjacent elements of the permutation.
\begin{align*}
 \pi &= \pi_1, \ldots , \pi_{a},\pi_{b}, \ldots ,\pi_m \\
  \pi^*&=  \pi_1, \ldots, \pi_{b},\pi_{a}, \ldots, \pi_m,
\end{align*}
with \added{$1\leq a < (m-1)$} and $b=a+1$.
\item \textbf{Interchange mutation}: Transposing two arbitrary elements of the permutation.
\begin{align*}
 \pi &= \pi_1, \ldots,\pi_{a-1}, \pi_{a},\pi_{a+1},\ldots,\pi_{b-1},\pi_{b},\pi_{b+1}, \ldots, \pi_m \\
  \pi^*&=   \pi_1, \ldots,\pi_{a-1}, \pi_{b},\pi_{a+1},\ldots,\pi_{b-1},\pi_{a},\pi_{b+1}, \ldots, \pi_m,
\end{align*}
with $1\leq a \leq m$ and $1\leq b \leq m$.
\item \textbf{Reversal mutation}: Reversing a substring of the permutation.
\begin{align*}
 \pi &= \pi_1, \ldots, \pi_{a}, \pi_{a+1},\ldots,\pi_{b-1}, \pi_{b}, \ldots, \pi_m \\
  \pi^*&=  \pi_1, \ldots, \pi_{b}, \pi_{b-1},\ldots,\pi_{a+1}, \pi_{a}, \ldots ,\pi_m ,
\end{align*}
with $1\leq a < b \leq m$.
\end{compactitem}

All 16 distance measures are
tested, with\\ $n= \{4 , \ldots , 20\} $ and  $m= \{4 , \ldots , 15\} $. With ten repeats, and the three different
submutation operators, this results into $97,920$ EA runs, each with $10,000$ fitness function evaluations.
The employed EA implementation is part of the R-package \texttt{CEGO}.

\subsection{Tests For Other Search Domains}
To show that the proposed approach is not limited to the presented permutation distance example, we
also briefly explore other search domains and their respective distances. \added{However, these are} not analyzed in further detail. Instead,
we provide a list of minimal examples in Appendix B: the smallest (w.r.t.\ dimension) indefinite distance matrix for each
tested distance measure. The examples include distance measures for permutations, signed permutations, 
trees, and strings.

\section{Observations and Discussion}\label{sec:results}
\subsection{Sampling Results}
\begin{figure}
\centering
\includegraphics[width=\linewidth]{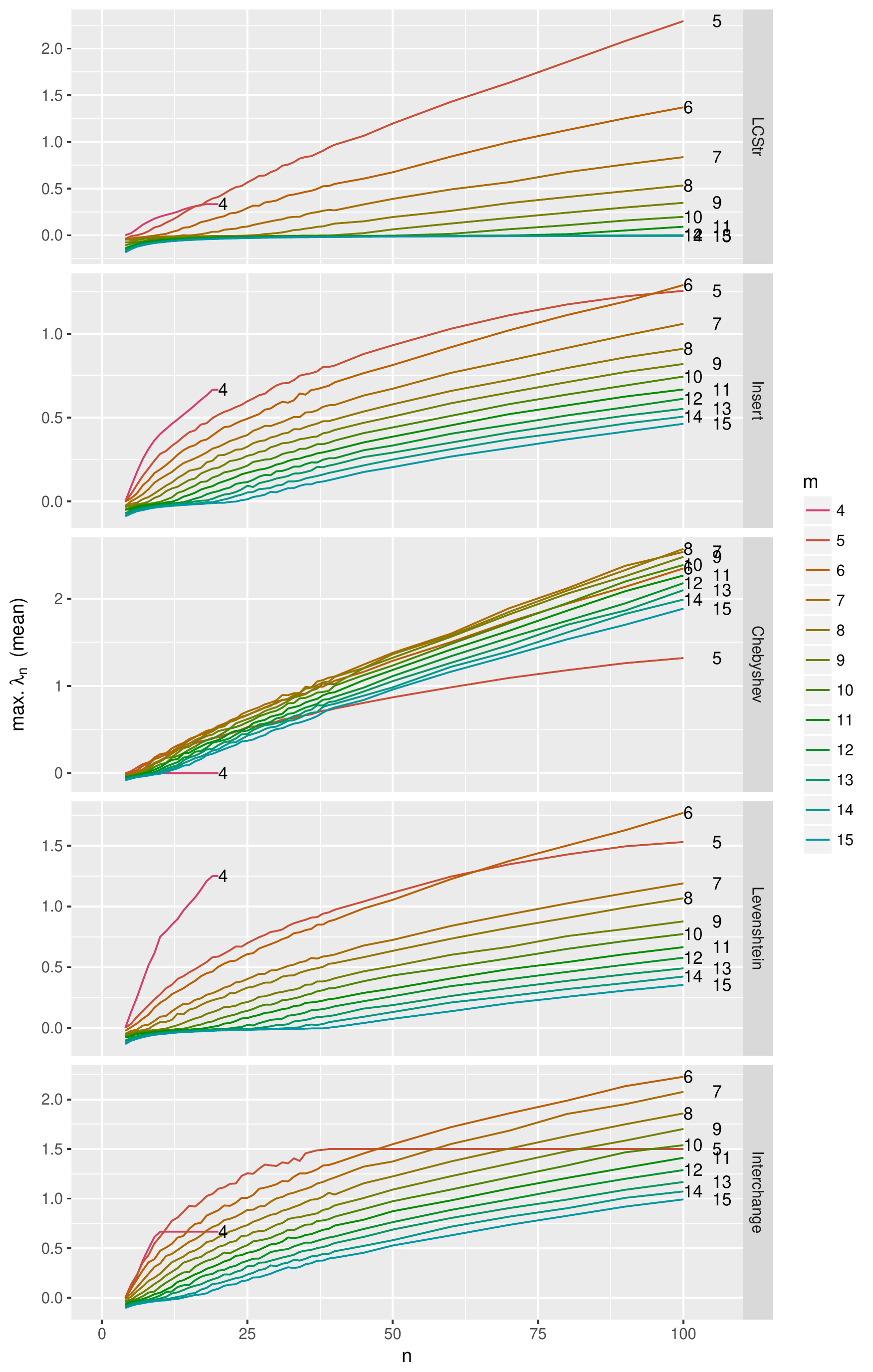}
\caption{Mean of maximum $\lambda_n$ \added{of $\hat{D}$} found with random sampling. The maximum is determined over all
sets in a single experiment. The mean is determined over ten repeats of the sampling. Only distance measures
\added{that} achieved positive $\lambda_n$ are shown. The numbers inside the plot and the color indicate the respective value of $m$.} \label{fig:meig} 
\end{figure}

The proportions of sets with positive eigenvalues (\peig) are summarized in Fig.~\ref{fig:peig}.
The largest eigenvalue\added{s are depicted} in Fig.~\ref{fig:meig}.
Only the five indefinite distance measures, which achieved positive eigenvalues are shown:
\begin{figure}
\centering
\includegraphics[width=\linewidth]{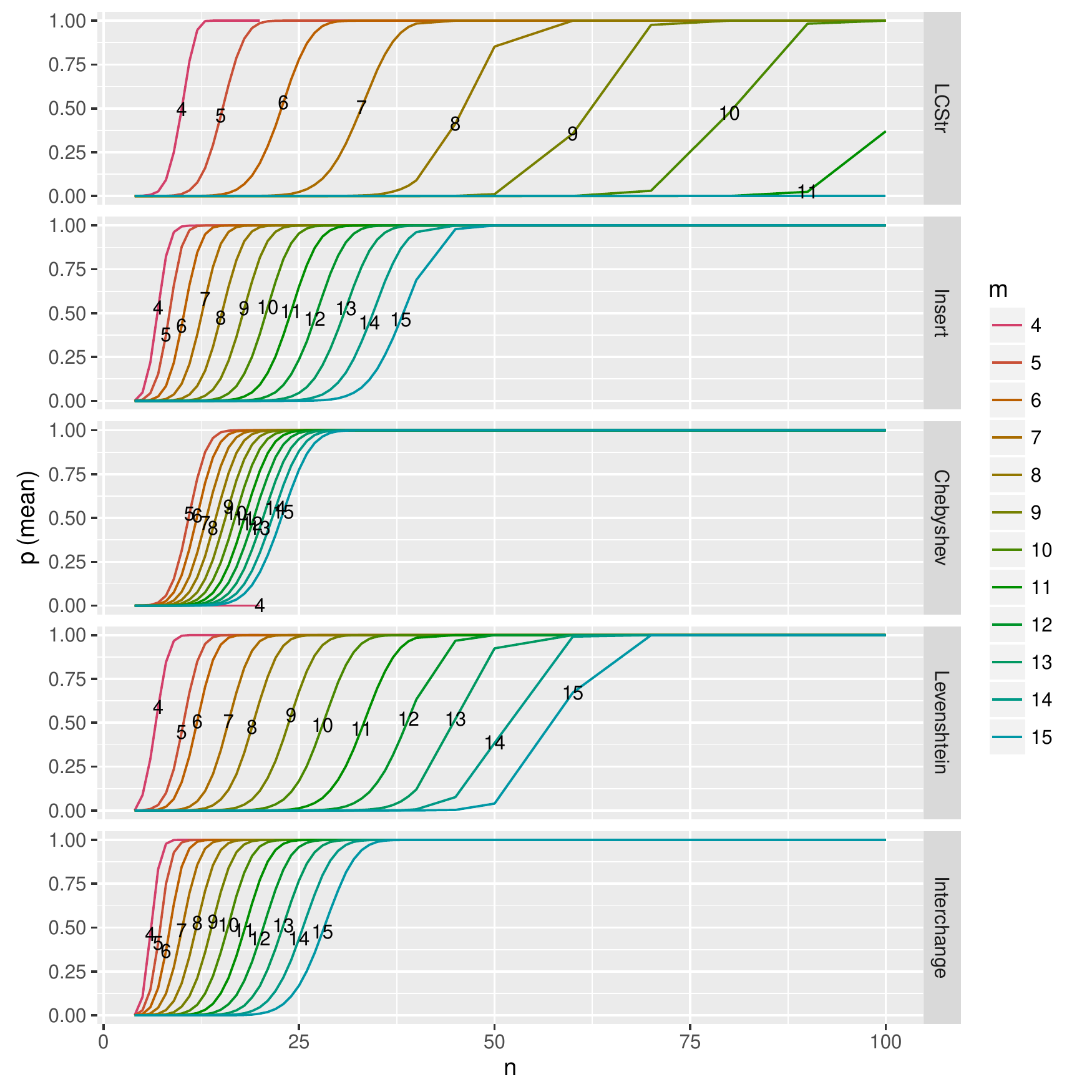}
\caption{Mean proportion of sets yielding matrices $\hat{D}$ with positive $\lambda_n$ found with random sampling. Only distance measures
\added{that} achieved positive $\lambda_n$ are shown. The numeric labels and the color indicate the respective value of $m$.} \label{fig:peig}
\end{figure} 
Longest Common Substring, Insert, Chebyshev, Levenshtein, Interchange.

No counter-examples are found for the remaining eleven measures. That does not
prove that \added{they} are CNSD (although some are CNSD, e.g., Euclidean distance, Swap 
distance~\citep{Jiao2015}, Hamming distance~\citep{Hutter2011}), 
but indicates that it may be unproblematic to use them in practice.
Some of the five non-CNSD distance measures were reported to work well
in a surrogate-model optimization framework, e.g., Levenshtein distance
seemed to work well for modeling of scheduling problems~\citep{Zaefferer2014c}.
This is mainly due to the fact, that \added{even a} non-CNSD distance \textit{may} yield
PSD kernel matrix $K$, depending on the specific data set and kernel parameters used.
We do not suggest that non-CNSD distance measures should be avoided, \added{but} that 
their application should be handled with care. 

Regarding \added{values of} \peig (Fig.~\ref{fig:peig}),
some trends can be observed. 
For \added{indefinite} distance measures, increasing the 
\added{set size} 
($n$) will in general lead to larger values of $\peig$. 
Obviously, a larger set is more likely to contain combinations of samples that yield
negative eigenvalues. In addition, the lower bound for eigenvalues decreases with increasing
matrix size~\citep{Constantine1985}.

In contrast to the set size, increasing the number of permutation elements ($m$) decreases the proportion of positive eigenvalues \peig in all five cases. This can be attributed to a larger and hence more difficult search space. 
Overall, none of the distance measures shows exactly the same behavior. LCStr distance
has the least problematic behavior. Only very few sets (of comparatively large size) yielded positive eigenvalues with \mbox{LCStr} distance.
Interchange, Levenshtein and Insert distance all have relatively large \peig for small set sizes $n$. Chebyshev on the other hand, 
starts to have non-zero \peig for relatively large $n$. However, the number of permutation elements has only a weak influence
in case of Chebyshev distance. Hence, curves for different $m$ are much closer to each other, compared to the other distance measures.
Somewhat analogous to $\peig$, the largest eigenvalues of $\hat{D}$ plotted in Fig.~\ref{fig:meig} are generally increasing for larger $n$, 
and decreasing with larger $m$.  

Our findings are confirmed by some results from literature.
\citet{Cortes2004a} have shown
that an exponential kernel function based on the Levenshtein distances 
is indefinite for strings of more than one symbol. Our experiments show that this result can be easily
rediscovered empirically, for the case of permutations.
At the same time, these findings also confirm (and provide reasons for)
problems observed with these kinds of kernel functions in a previous study~\citep{Zaefferer2014c}.

As a consistency check, Table~\ref{tab:brute} compares the sampling
results to a brute force approach for $n= \{4 , \ldots , 8\} $ and $m=4$.
It shows that the sampling indeed approximates
the number of non-CNSD matrices quite well. In this small set, the sampling identified all combinations of distance measure and $n$ that may yield non-CNSD matrices.
\begin{table}[ht]
\centering
\caption{Comparing the true proportion of non-CNSD matrices \peig determined by brute force (true) with the mean of \peig estimated by sampling (estimate).
The table only presents results for permutations with $m=4$ elements and set sizes $n$. 
The number of sets evaluated by brute force is $n_s=n_p!  \ / \  (n! \ (n_p-n)! )$ with $n_p=m!$.
}\label{tab:brute}
\begin{tabular}{rlrcc}
  \toprule
 n & distance & \multicolumn{1}{c}{$n_s$} & true & estimate \\ 
  \midrule
 5 & Insert & 42,504 & 0.047 & 0.048 \\ 
 6 & Insert & 134,596 & 0.221 & 0.219 \\ 
 7 & Insert & 346,104 & 0.530 & 0.529 \\ 
  8 & Insert & 735,471 & 0.827 & 0.825 \\ 
	\midrule
  5 & Interchange & 42,504 & 0.106 & 0.105 \\ 
  6 & Interchange & 134,596 & 0.465 & 0.463 \\ 
 7 & Interchange & 346,104 & 0.833 & 0.834 \\ 
  8 & Interchange & 735,471 & 0.978 & 0.978 \\ 
	\midrule
  5 & Levenshtein & 42,504 & 0.087 & 0.087 \\ 
  6 & Levenshtein & 134,596 & 0.293 & 0.293 \\ 
  7 & Levenshtein & 346,104 & 0.591 & 0.589 \\ 
  8 & Levenshtein & 735,471 & 0.847 & 0.846 \\ 
	\midrule
  5 & LCStr & 42,504 & 0.002 & 0.002 \\ 
  6 & LCStr & 134,596 & 0.007 & 0.007 \\ 
  7 & LCStr & 346,104 & 0.026 & 0.026 \\ 
  8 & LCStr & 735,471 & 0.093 & 0.092 \\ 
   \bottomrule
\end{tabular}
\end{table}

\subsection{Optimization Results}
\begin{figure*}
\centering
\includegraphics[width=\linewidth]{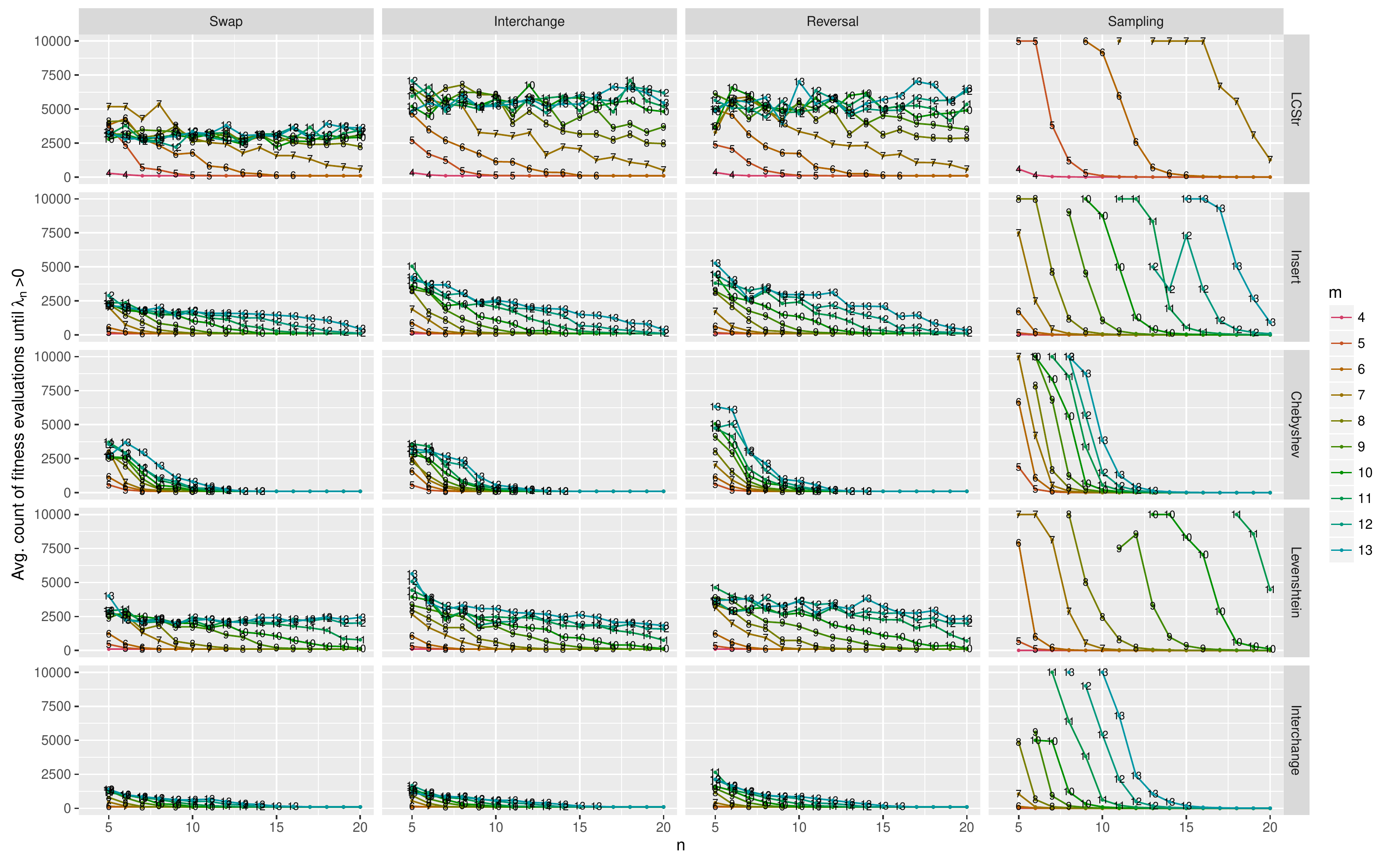}
\caption{Average number of fitness function evaluations until a positive eigenvalue was found. 
Missing nodes indicate that no positive eigenvalues are found, within the given budget. Averages are computed after removal of cases where no positive values are found.
Columns show the results of the EA with each submutation function and the earlier described sampling approach.
The rows indicate different distances measures.
Only distance measures which achieved positive eigenvalues are shown.
 } \label{fig:teig}
\end{figure*}

The average number of fitness function evaluations (i.e., number of times $\lambda_n$ \added{of $\hat{D}$} is computed)
required to find a matrix $\hat{D}$ with positive $\lambda_n$ is depicted in Fig.~\ref{fig:teig}.
The optimization results show similar behavior with respect to $n$ and $m$ as the sampling approach. Increasing $m$
leads to an increased number of fitness function evaluations. That means, finding positive eigenvalues becomes more difficult with increasing
values of $m$.
Increasing $n$ reduces the number of required fitness function evaluations. That is, finding positive eigenvalues becomes \added{easier}.
In some cases, this effect disappears for large values of $m$, e.g., for LCStr distance, where the average is more or less constant over
$n$, if $m$ is large enough. 

Importantly, the comparison to the sampling results clearly shows that the EA has some success in optimizing the largest eigenvalue
of the transformed distance matrix $\hat{D}$. 
In several cases, positive eigenvalues are found by the EA while sampling with the same budget
failed to find any. 
\added{Hence}, it can be assumed that \added{the fitness landscape based on $\lambda_n$ \added{of $\hat{D}$}} is sufficiently smooth to allow for optimization.
 \added{The eigenvalue $\lambda_n$ seems to be a good} indicator of how close a solution set is to yielding a\added{n} indefinite kernel matrix.

Furthermore, the expected bias of the used submutation operator becomes visible. 
For Insert and LCStr distance, the EA with swap mutation works considerably better than the other two variants.
Hence, comparisons of these values across different distance measures should \added{be} handled with caution.
Clearly, other aspects of the optimization algorithm (e.g., selection criteria or recombination operators) might have similar effects.
While this bias is troubling, results may still offer interesting insights. The eigenvalue optimization
could be interpreted as a worst-case scenario that occurs if an iterative learning process strongly correlates with the 
eigenvalues of the employed distance matrix.

\subsection{Verification: Impact on Model Quality}\label{sec:modelQuality}
\added{
Earlier, we discussed two values 
that may express the effect of the lack of definiteness
in practice, i.e., $\peig$ and $\lambda_n$ of $\hat{D}$.
But what do these values imply?
}

\added{
The value $\peig$ can be seen as a probability of generating indefinite matrices.
If we assume that a model is unable to deal with indefinite data,
the fraction $\peig$ is an estimate of how likely a modeling failure is. 
For other cases, it is very hard to link it to a model performance measure
such as accuracy without making too many assumptions.
We still argue that $\peig$ provides useful information, especially
when the kernel is designed and probed before sampling any data (e.g., 
when planning an experiment).
We suggest to use $\peig$ to support an initial decision (e.g., whether
to spend additional time on fixing or otherwise dealing with the indefinite kernel).
One advantage is that it is rather easy to interpret.
}

\added{
In contrast, the parameter $\lambda_n$ of $\hat{D}$ is more difficult to interpret. 
But it has the advantage that it may be estimated for a single matrix
as well as its average for a set of matrices. Hence, we want to determine
 whether the magnitude of this eigenvalue affects model performance. 
We expect an influence that depends on the choice of model.
Consider, e.g., a Gaussian process regression model, as e.g., described by \cite{Forrester2008a}.
The model may be able to mitigate the problematic eigenvalue
by assigning larger $\theta$ values to the kernel $k(x,x')=\exp(-\theta \fdist(x,x'))$. 
For very large $\lambda_n$ and thus very large $\theta$, this will lead to kernel matrices
that approximate the unit matrix, which is positive definite.
A model with a unit kernel matrix would be able to reproduce
the training data, but would predict the process mean for most other data points.
Hence, we examine Gaussian process regression models, since they provide
a transparent and interpretable test case (but similar experiments
could easily be made with support vector regression).
}

\added{
An experimental test has to consider the potential bias of the used data set.
We need to be able to reasonably assume that differences in performance
are actually due to the properties of employed distance or kernel 
(i.e., a kernel performs poorly because the corresponding $\lambda_n$ of $\hat{D}$ is high) 
rather than properties of the data set
(i.e., a kernel performs poorly because it does not fit well to the ground
truth of the data set).
To that end, we suggest that observations in a test data set are derived from
the same distances that are used in the model.
}

\added{
Hence, we randomly created data sets $X$ of size $n$ with permutations of dimension $m$,
similarly to the random sampling performed earlier.
Then, we created training observations by evaluating the distance of each permutation in $X$
to a reference permutation $x_{ref}={1,...,m}$, i.e., $y=\fdist(x,x_{ref})$. 
A Gaussian process model was then trained with this data, largely following
the descriptions in~\cite{Forrester2008a}. 
The model was trained with the kernel $k(x,x')=\exp(-\theta \fdist(x,x'))$,
and the model parameters (e.g., $\theta$) were determined by maximum likelihood estimation, via the locally biased
version of the DIviding RECTangles (DIRECT) algorithm \citep{Gablonsky2001}
with 1,000 likelihood evaluations.
For each test, the distance chosen to produce the observations $y$ and the distance chosen in the kernel were identical.
}

\begin{figure}
\centering
\includegraphics[width=\linewidth]{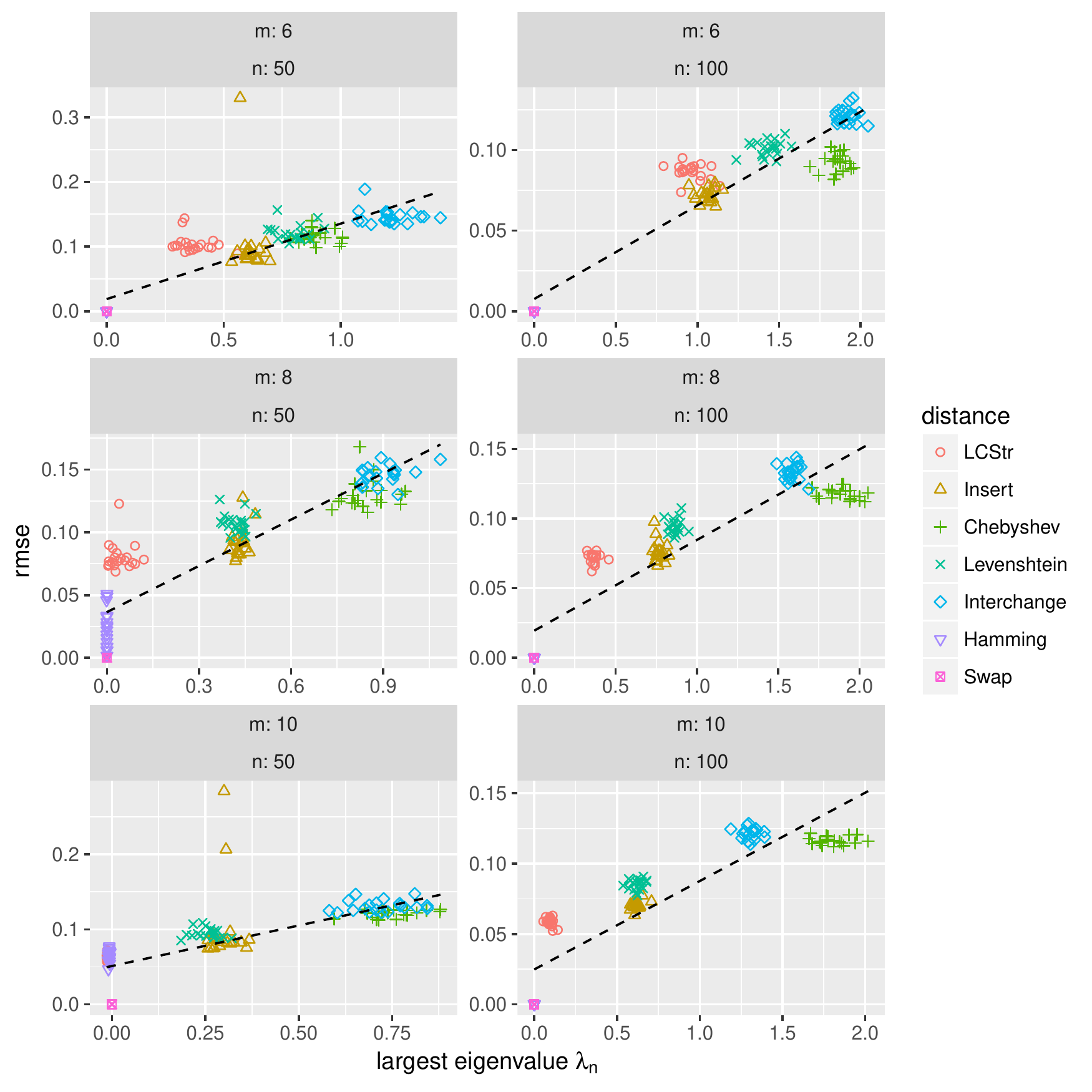}
\caption{\added{RMSEs of a Gaussian process regression model plotted against the eigenvalue $\lambda_n$ of $\hat{D}$ that is critical to definiteness of the underlying distance matrices. The denoted distance measures are used in the underlying data set
as well as the model itself. The dashed line depicts a linear trend.}} \label{fig:eig1}
\end{figure}

\added{
The Root Mean
Square Error (RMSE)  of the model was evaluated on 1000 randomly chosen permutations.
The resulting RMSE values for each training set $X$, as well as the corresponding
eigenvalue $\lambda_n$ of $\hat{D}$ are shown in Fig.~\ref{fig:eig1}.
The graphic shows a trend that confirms our expectation. It seems that distances
associated to larger $\lambda_n$ tend to produce larger errors.
}
%
\section{Summary and Outlook}\label{sec:summary}
The focus of this study was the definiteness of kernel and distance measures.
Definiteness is a main requirement for modeling techniques like Gaussian processes or SVMs.
Distance-based kernels with unknown definiteness
may be promising choices for certain parameter spaces.
Yet, their definiteness may be very hard to prove or enforce.

It was found, that empirical approaches (based on sampling or optimization) may help
to assess definiteness of the respective function. In detail, two research questions
were investigated:

\begin{description}
\item[\Qone] \textbf{Discovery:} Is there an efficient, empirical approach to determine the 
definiteness of kernel functions based on arbitrary distance measures? 
\item[\Qtwo] \textbf{Measurability:}  If \Qone can be answered affirmatively, 
\added{can we quantify to what extent a lack of definiteness is problematic?}
\end{description}

Two empirical approaches were suggested towards that end. The first approach (\ASone)
samples from the space of solution sets, and determines \added{whether} a set is found which leads
to an indefinite distance or kernel matrix. If indefinite matrices are rare, a directed search with an EA
is more successful (\AStwo). The EA maximizes the largest eigenvalue of a transformed distance matrix $\hat{D}$, 
respectively minimizes the smallest eigenvalue of a kernel matrix. Hence, the EA searches for sets \added{that} yield indefinite matrices. 

As a proof-of-concept, the approaches were applied to distance measures for permutations. 
It was shown that five problematic distance measures could
be identified: Longest Common Substring (LCStr), Insert, Chebyshev, Levenshtein, and Interchange distance.
Information known from literature (regarding indefiniteness of the respective
kernel function) could be rediscovered by the empirical approaches.

The optimization approach was successful, as it was able to outperform the sampling approach
in discovering sets with indefinite kernel matrices. Still, the results also indicated that the
choice of variation operators in the optimization algorithm does introduce bias. 
Hence, the respective results do not allow a conclusion about the \added{impact} of \added{a lack of} definiteness
of the respective sets/matrices. 
Still, the success of the EA indicates that the fitness landscape posed by the largest eigenvalue is
not excessively rugged and has an exploitable structure. This suggests that the largest eigenvalue is a good indicator of how far a
certain solution set (and the respective distance or kernel matrix) is from becoming indefinite.
\added{In an additional set of experiments, we further verified that increasing the largest eigenvalue can in fact
be linked to a decrease in model quality.}
This results into the following responses to the posed research questions:

\begin{description}
\item[R1] \textbf{Discovery:} Sampling from the space of potential candidate solution sets allows identifying problems with definiteness,
by identifying solution sets that lead to non-CNSD distance matrices (\ASone). 
Where such situations are rare (hence more likely to be missed by the sampling),
an optimization approach may \added{be more successful} (\AStwo).
While neither approach \ASone nor \AStwo are able to prove definiteness, both are able to disprove it. 
If no negative results are found it is reasonable to assume that 
using the respective distance/kernel function is feasible.
\item[R2] \textbf{Measurability:} The sampling approach (\ASone) yields a proportion of potentially non-CNSD matrices, which in turn yields an estimate
of how problematic a distance measure is. In a similar way, yet potentially biased by the choice of optimization algorithm,
the number of evaluations required by the optimization approach gives a similar estimate.
In addition, the success of the optimization approach (\AStwo) suggests that the respective largest eigenvalue 
is an indicator of how close certain sets (and respective distance or kernel matrices) are to becoming indefinite.
\added{Additional experiments showed how this eigenvalue could be linked to model performance.}
\end{description}

For future research, it may be of interest to allow the EA to change the set size. 
Clearly, one issue would be that enlarging the sets may quickly lead to a trivial solution, since
larger sets naturally lead to larger $\lambda_n$ \added{of $\hat{D}$}.
Hence, there is a trade-off between \added{the} largest eigenvalue and \added{the} set size.
A multi-objective EA (e.g. NSGA-II~\citep{Deb2002} or SMS-EMOA~\citep{Beume2007}) 
may be used to handle this issue by simultaneously 
maximizing $\lambda_n$ and minimizing the set size $n$.

Finally, the herein described kernels and distances are not the full story. For other kernels, the relation between distance measure and kernel function
may not be as straightforward. Parameters of the distance measure or the kernel function 
could complicate the situation. It may be necessary to adapt the proposed method to, e.g., include
parameters in the sampling and optimization procedures. 

\medskip

\noindent
\textbf{Compliance with ethical standards}\\

\noindent
\textbf{Conflicts of Interest~~} The authors declare that they have no conflict of interest.


\bibliographystyle{myapalike} 
\bibliography{Zaef14e}

\begin{thebibliography}{}

\bibitem[Bader et~al., 2004]{Bader2004a}
Bader, D.~A., Moret, B.~M., Warnow, T., Wyman, S.~K., Yan, M., Tang, J.,
  Siepel, A.~C., \& Caprara, A. (2004).
\newblock Genome rearrangements analysis under parsimony and other phylogenetic
  algorithms (grappa) 2.0.
\newblock Online: https://www.cs.unm.edu/~moret/GRAPPA/.
\newblock accessed: 16.11.2016.

\bibitem[Bartz-Beielstein and Zaefferer, 2017]{Bartz-Beielstein2016n}
Bartz-Beielstein, T. \& Zaefferer, M. (2017).
\newblock Model-based methods for continuous and discrete global optimization.
\newblock {\em Applied Soft Computing}, 55:154 -- 167.

\bibitem[Berg et~al., 1984]{Berg1984}
Berg, C., Christensen, J. P.~R., \& Ressel, P. (1984).
\newblock {\em Harmonic Analysis on Semigroups}, volume 100 of {\em Graduate
  Texts in Mathematics}.
\newblock Springer New York.

\bibitem[Beume et~al., 2007]{Beume2007}
Beume, N., Naujoks, B., \& Emmerich, M. (2007).
\newblock {SMS}-{EMOA}: Multiobjective selection based on dominated
  hypervolume.
\newblock {\em European Journal of Operational Research}, 181(3):1653--1669.

\bibitem[Boytsov, 2011]{Boytsov2011}
Boytsov, L. (2011).
\newblock Indexing methods for approximate dictionary searching: Comparative
  analysis.
\newblock {\em J. Exp. Algorithmics}, 16:1--91.

\bibitem[Burges, 1998]{Burges1998}
Burges, C.~J. (1998).
\newblock A tutorial on support vector machines for pattern recognition.
\newblock {\em {Data Mining and Knowledge Discovery}}, 2(2):121--167.

\bibitem[Camastra and Vinciarelli, 2008]{Camastra2007}
Camastra, F. \& Vinciarelli, A. (2008).
\newblock {\em {Machine Learning for Audio, Image and Video Analysis: Theory
  and Applications}}.
\newblock Advanced information and knowledge processing. Springer, London.

\bibitem[Campos et~al., 2005]{Campos2005}
Campos, V., Laguna, M., \& Mart{\'\i}, R. (2005).
\newblock Context-independent scatter and tabu search for permutation problems.
\newblock {\em INFORMS Journal on Computing}, 17(1):111--122.

\bibitem[Camps-Valls et~al., 2004]{Camps-Valls2004}
Camps-Valls, G., Mart{\'\i}n-Guerrero, J.~D., Rojo-{\'A}lvarez, J.~L., \&
  Soria-Olivas, E. (2004).
\newblock Fuzzy sigmoid kernel for support vector classifiers.
\newblock {\em Neurocomputing}, 62:501--506.

\bibitem[Chen et~al., 2009]{Chen2009}
Chen, Y., Gupta, M.~R., \& Recht, B. (2009).
\newblock Learning kernels from indefinite similarities.
\newblock In {\em Proceedings of the 26th Annual International Conference on
  Machine Learning}, ICML '09,  (pp. 145--152), New York, NY, USA. ACM.

\bibitem[Constantine, 1985]{Constantine1985}
Constantine, G. (1985).
\newblock Lower bounds on the spectra of symmetric matrices with nonnegative
  entries.
\newblock {\em Linear Algebra and its Applications}, 65:171--178.

\bibitem[Cortes et~al., 2004]{Cortes2004a}
Cortes, C., Haffner, P., \& Mohri, M. (2004).
\newblock Rational kernels: Theory and algorithms.
\newblock {\em J. Mach. Learn. Res.}, 5:1035--1062.

\bibitem[Curriero, 2006]{Curriero2006}
Curriero, F. (2006).
\newblock On the use of non-euclidean distance measures in geostatistics.
\newblock {\em Mathematical Geology}, 38(8):907--926.

\bibitem[Deb et~al., 2002]{Deb2002}
Deb, K., Pratap, A., Agarwal, S., \& Meyarivan, T. (2002).
\newblock A fast and elitist multiobjective genetic algorithm: {NSGA}-{II}.
\newblock {\em {IEEE} Transactions on Evolutionary Computation}, 6(2):182--197.

\bibitem[Deza and Huang, 1998]{Deza1998}
Deza, M. \& Huang, T. (1998).
\newblock Metrics on permutations, a survey.
\newblock {\em Journal of Combinatorics, Information and System Sciences},
  23(1-4):173--185.

\bibitem[Eiben and Smith, 2003]{Eiben2003}
Eiben, A.~E. \& Smith, J.~E. (2003).
\newblock {\em Introduction to Evolutionary Computing}.
\newblock Springer Berlin Heidelberg.

\bibitem[Feller, 1971]{Feller1971}
Feller, W. (1971).
\newblock {\em An Introduction to Probability Theory and Its Applications,
  Volume 2}.
\newblock JOHN WILEY \& SONS INC.

\bibitem[Forrester et~al., 2008]{Forrester2008a}
Forrester, A., Sobester, A., \& Keane, A. (2008).
\newblock {\em Engineering Design via Surrogate Modelling}.
\newblock Wiley.

\bibitem[Gablonsky and Kelley, 2001]{Gablonsky2001}
Gablonsky, J. \& Kelley, C. (2001).
\newblock A locally-biased form of the direct algorithm.
\newblock {\em Journal of Global Optimization}, 21(1):27--37.

\bibitem[G\"{a}rtner et~al., 2003]{Gaertner2003a}
G\"{a}rtner, T., Lloyd, J., \& Flach, P. (2003).
\newblock Kernels for structured data.
\newblock In Matwin, S. \& Sammut, C., editors, {\em Inductive Logic
  Programming}, volume 2583 of {\em Lecture Notes in Computer Science}, pages
  66--83. Springer Berlin Heidelberg.

\bibitem[G\"{a}rtner et~al., 2004]{Gaertner2004}
G\"{a}rtner, T., Lloyd, J., \& Flach, P. (2004).
\newblock Kernels and distances for structured data.
\newblock {\em Machine Learning}, 57(3):205--232.

\bibitem[Haussler, 1999]{Haussler1999}
Haussler, D. (1999).
\newblock Convolution kernels on discrete structures.
\newblock Technical Report UCSC-CRL-99-10, Department of Computer Science,
  University of California at Santa Cruz.

\bibitem[Hirschberg, 1975]{Hirschberg1975}
Hirschberg, D.~S. (1975).
\newblock A linear space algorithm for computing maximal common subsequences.
\newblock {\em Commun. ACM}, 18(6):341--343.

\bibitem[Hutter et~al., 2011]{Hutter2011}
Hutter, F., Hoos, H.~H., \& Leyton-Brown, K. (2011).
\newblock Sequential model-based optimization for general algorithm
  configuration.
\newblock In {\em Proc.~of LION-5},  (pp. 507--523).

\bibitem[Ikramov and Savel'eva, 2000]{Ikramov2000}
Ikramov, K. \& Savel'eva, N. (2000).
\newblock Conditionally definite matrices.
\newblock {\em Journal of Mathematical Sciences}, 98(1):1--50.

\bibitem[Jiao and Vert, 2015]{Jiao2015}
Jiao, Y. \& Vert, J.-P. (2015).
\newblock The {Kendall} and {Mallows} kernels for permutations.
\newblock In {\em Proceedings of the 32nd International Conference on Machine
  Learning (ICML-15)},  (pp. 1935--1944).

\bibitem[Kendall and Gibbons, 1990]{Kendall1990}
Kendall, M. \& Gibbons, J. (1990).
\newblock {\em {Rank Correlation Methods}}.
\newblock Oxford University Press.

\bibitem[Lee, 1958]{Lee1958}
Lee, C. (1958).
\newblock Some properties of nonbinary error-correcting codes.
\newblock {\em IRE Transactions on Information Theory}, 4(2):77--82.

\bibitem[Li and Jiang, 2004]{Li2004}
Li, H. \& Jiang, T. (2004).
\newblock A class of edit kernels for svms to predict translation initiation
  sites in eukaryotic mrnas.
\newblock In {\em Proceedings of the Eighth Annual International Conference on
  Resaerch in Computational Molecular Biology}, RECOMB '04,  (pp. 262--271),
  New York, NY, USA. ACM.

\bibitem[Loosli et~al., 2015]{Loosli2015}
Loosli, G., Canu, S., \& Ong, C. (2015).
\newblock Learning {SVM} in {Krein} spaces.
\newblock {\em IEEE Transactions on Pattern Analysis and Machine Intelligence},
  38(6):1204--1216.

\bibitem[Marteau and Gibet, 2014]{Marteau2014}
Marteau, P.-F. \& Gibet, S. (2014).
\newblock On recursive edit distance kernels with application to time series
  classification.
\newblock {\em IEEE Transactions on Neural Networks and Learning Systems},
  PP(99):1--1.

\bibitem[Moraglio and Kattan, 2011]{Moraglio2011}
Moraglio, A. \& Kattan, A. (2011).
\newblock Geometric generalisation of surrogate model based optimisation to
  combinatorial spaces.
\newblock In {\em Proceedings of the 11th European Conference on Evolutionary
  Computation in Combinatorial Optimization}, EvoCOP'11,  (pp. 142--154),
  Berlin, Heidelberg, Germany. Springer.

\bibitem[Motwani and Raghavan, 1995]{MR95}
Motwani, R. \& Raghavan, P. (1995).
\newblock {\em Randomized Algorithms}.
\newblock Cambridge University Press.

\bibitem[Murphy, 2012]{Murphy2012}
Murphy, K.~P. (2012).
\newblock {\em Machine Learning}.
\newblock MIT Press Ltd.

\bibitem[Ong et~al., 2004]{Ong2004}
Ong, C.~S., Mary, X., Canu, S., \& Smola, A.~J. (2004).
\newblock Learning with non-positive kernels.
\newblock In {\em Proceedings of the Twenty-first International Conference on
  Machine Learning}, ICML '04,  (pp. 81--88), New York, NY, USA. ACM.

\bibitem[Pawlik and Augsten, 2015]{Pawlik2015}
Pawlik, M. \& Augsten, N. (2015).
\newblock Efficient computation of the tree edit distance.
\newblock {\em {ACM} Transactions on Database Systems}, 40(1):1--40.

\bibitem[Pawlik and Augsten, 2016]{Pawlik2016}
Pawlik, M. \& Augsten, N. (2016).
\newblock Tree edit distance: Robust and memory-efficient.
\newblock {\em Information Systems}, 56:157--173.

\bibitem[Rasmussen and Williams, 2006]{Rasmussen2006}
Rasmussen, C.~E. \& Williams, C. K.~I. (2006).
\newblock {\em Gaussian Processes for Machine Learning}.
\newblock The MIT Press.

\bibitem[Reeves, 1999]{Reeves1999}
Reeves, C.~R. (1999).
\newblock Landscapes, operators and heuristic search.
\newblock {\em Annals of Operations Research}, 86(0):473--490.

\bibitem[Schiavinotto and St{\"u}tzle, 2007]{Schiavinotto2007}
Schiavinotto, T. \& St{\"u}tzle, T. (2007).
\newblock A review of metrics on permutations for search landscape analysis.
\newblock {\em {Computers \& Operations Research}}, 34(10):3143--3153.

\bibitem[Schleif and Tino, 2015]{Schleif2015}
Schleif, F.-M. \& Tino, P. (2015).
\newblock Indefinite proximity learning: A review.
\newblock {\em Neural Computation}, 27(10):2039--2096.

\bibitem[Sch\"{o}lkopf, 2001]{Schoelkopf2001}
Sch\"{o}lkopf, B. (2001).
\newblock The kernel trick for distances.
\newblock In Leen, T.~K., Dietterich, T.~G., \& Tresp, V., editors, {\em
  Advances in Neural Information Processing Systems 13}, pages 301--307. MIT
  Press.

\bibitem[Sevaux and S\"{o}rensen, 2005]{Sevaux2005}
Sevaux, M. \& S\"{o}rensen, K. (2005).
\newblock Permutation distance measures for memetic algorithms with population
  management.
\newblock In {\em Proceedings of 6th Metaheuristics International Conference
  (MIC’05)},  (pp. 832--838). University of Vienna.

\bibitem[Singhal, 2001]{Singhal2001}
Singhal, A. (2001).
\newblock Modern information retrieval: a brief overview.
\newblock {\em IEEE Bulletin on Data Engineering}, 24(4):35--43.

\bibitem[Smola et~al., 2000]{Smola2000}
Smola, A.~J., Ovári, Z.~L., \& Williamson, R.~C. (2000).
\newblock Regularization with dot-product kernels.
\newblock In {\em Advances in Neural Information Processing Systems 13,
  Proceedings},  (pp. 308--314). MIT Press.

\bibitem[van~der Loo, 2014]{Loo2014}
van~der Loo, M.~P. (2014).
\newblock The stringdist package for approximate string matching.
\newblock {\em The R Journal}, 6(1):111--122.

\bibitem[Vapnik, 1998]{Vapnik1998}
Vapnik, V.~N. (1998).
\newblock {\em Statistical Learning Theory}, volume~1.
\newblock Wiley New York.

\bibitem[Voutchkov et~al., 2005]{Voutchkov2005}
Voutchkov, I., Keane, A., Bhaskar, A., \& Olsen, T.~M. (2005).
\newblock Weld sequence optimization: The use of surrogate models for solving
  sequential combinatorial problems.
\newblock {\em Computer Methods in Applied Mechanics and Engineering},
  194(30-33):3535--3551.

\bibitem[Wagner and Fischer, 1974]{Wagner1974}
Wagner, R.~A. \& Fischer, M.~J. (1974).
\newblock The string-to-string correction problem.
\newblock {\em J. ACM}, 21(1):168--173.

\bibitem[Wu et~al., 2005]{Wu2005}
Wu, G., Chang, E.~Y., \& Zhang, Z. (2005).
\newblock An analysis of transformation on non-positive semidefinite similarity
  matrix for kernel machines.
\newblock In {\em Proceedings of the 22nd International Conference on Machine
  Learning}.

\bibitem[Zaefferer and Bartz-Beielstein, 2016]{Zaefferer2016b}
Zaefferer, M. \& Bartz-Beielstein, T. (2016).
\newblock Efficient global optimization with indefinite kernels.
\newblock In {\em Parallel Problem Solving from Nature--PPSN XIV},  (pp.
  69--79). Springer.

\bibitem[Zaefferer et~al., 2014a]{Zaefferer2014c}
Zaefferer, M., Stork, J., \& Bartz-Beielstein, T. (2014a).
\newblock Distance measures for permutations in combinatorial efficient global
  optimization.
\newblock In Bartz-Beielstein, T., Branke, J., Filipi{\v c}, B., \& Smith, J.,
  editors, {\em Parallel Problem Solving from Nature--PPSN XIII},  (pp.
  373--383), Cham, Switzerland. Springer.

\bibitem[Zaefferer et~al., 2014b]{Zaefferer2014b}
Zaefferer, M., Stork, J., Friese, M., Fischbach, A., Naujoks, B., \&
  Bartz-Beielstein, T. (2014b).
\newblock Efficient global optimization for combinatorial problems.
\newblock In {\em Proceedings of the 2014 Conference on Genetic and
  Evolutionary Computation}, GECCO '14,  (pp. 871--878), New York, NY, USA.
  ACM.

\end{thebibliography}

\appendix

\section*{Appendix A: Distance Measures for Permutations}

In the following, we describe the distance measures employed in the experiments.

\begin{itemize}[leftmargin=0em,itemindent=*,topsep=0cm]
\item The Levenshtein distance is an edit distance measure: \\ \smallskip
 \centerline{ $\fdist_{Lev}(\pi,\pi')  = edits_{\pi\rightarrow \pi'}$}
Here, $edits_{\pi\rightarrow \pi'}$ is the minimal number of deletions, insertions, or substitutions
required to transform one string
(or here: permutation) $\pi$ into another string $\pi'$. 
The implementation is based on \citet{Wagner1974}. 
\item Swaps are transpositions of two adjacent elements.
The Swap distance (also: Kendall's Tau \citep{Kendall1990,Sevaux2005} or
Precedence distance \citep{Schiavinotto2007}) 
counts the minimum number of swaps  
required to transform one permutation into another. 
 For permutations, it is~\citep{Sevaux2005}: 
\begin{align*}
\fdist_{Swa}(\pi,\pi')  = \sum_{i=1}^{m} \sum_{j=1}^{m} z_{ij} ~~  \text{with}\\
z_{ij} = \left\{
  \begin{array}{l l}
    1 & \quad \text{if }  \pi_i < \pi_j ~\text{and}~ \pi'_i > \pi'_j ,\\
    0 & \quad \text{otherwise.}
  \end{array} \right.
\end{align*}
\item An interchange operation is the transposition of two arbitrary elements. 
Respectively, the Interchange (also: Cayley) distance counts the minimum number
 of interchanges ($interchanges_{\pi\rightarrow \pi'}$) required to transform one permutation into another \citep{Schiavinotto2007}:
 \smallskip \\ \smallskip
 \centerline{ $\fdist_{Int}(\pi,\pi')  = interchanges_{\pi\rightarrow \pi'}$
}
\item The Insert distance is based on the longest common subsequence $LCSeq(\pi,\pi')$.
The longest common subsequence is the largest number of elements that follow each 
other in both permutations, with interruptions. The corresponding
distance is \\ \smallskip
\centerline{$\fdist_{Ins}(\pi,\pi')  = m-LCSeq(\pi,\pi').$}
We use the algorithm described by \citet{Hirschberg1975}.
The name is due to its interpretation as an edit distance measure. The corresponding edit operation is a combination of insertion
and deletion. A single element is moved from one position (delete) to a new position (insert).
It is also called Ulam's distance \citep{Schiavinotto2007}. 
\item The Longest Common Substring distance is based on the largest number of 
elements that follow each other in both permutations, without interruption.
Unlike the longest common subsequence all elements have to be adjacent. 

If $LCStr(\pi,\pi')$ is the length of the longest common string,
the distance is
\begin{equation*}
 \fdist_{LCStr}(\pi,\pi')= m-LCStr(\pi,\pi').
\end{equation*} 
\item  The R-distance~\citep{Campos2005,Sevaux2005} 
counts the number of times that one element follows another in one permutation, but not in the other. 
It is identical with the uni-directional adjacency distance~\citep{Reeves1999}. It is computed by  
\begin{align*}
\fdist_{R}(\pi,\pi')  = \sum_{i=1}^{m-1} y_i ~~ \text{with}~~~~~~~~~~~~~~~~~\\
 y_i = \left\{
  \begin{array}{l l}
   0 & \quad \text{if }\exists j : \pi_i=\pi'_j ~\text{and}~ \pi_{i+1}=\pi'_{j+1} ,\\
   1 & \quad \text{otherwise.}
  \end{array} \right.
\end{align*}
\item The (bi-directional) Adjacency 
distance~\citep{Reeves1999,Schiavinotto2007} counts the number of times two elements are neighbors in one, but not in the other permutation.
Unlike R-distance (uni-directional), the order of the two elements does not matter.
It is computed by 
\begin{align*}
\fdist_{Adj}(\pi,\pi')  = \sum_{i=1}^{m-1} y_i ~~ \text{with}~~~~~~~~~~~~~~~~~~~~~~~~~~\\
 y_i = \left\{
  \begin{array}{l l}
   0 & \quad \text{if }\exists j : \pi_i=\pi'_j ~\text{and}~ \pi_{i+1} \in \{\pi'_{j+1}, \pi'_{j-1} \},\\
   1 & \quad \text{otherwise.}
  \end{array} \right.
\end{align*}

\item The Position distance~\citep{Schiavinotto2007} is identical with the Deviation distance or Spearman's footrule \citep{Sevaux2005}, \smallskip \\ \smallskip
 \centerline{ $\fdist_{\text{Pos}}(\pi,\pi')  = \sum_{k=1}^{m} |i-j | ~~\text{where}~~\pi_i = \pi'_j = k$ .}
\item The non-metric Squared Position distance is Spearman's rank correlation coefficient~\citep{Sevaux2005}. In contrast to the Position distance, the term $|i-j|$ is replaced by  $(i-j)^2$.
\item The Hamming distance or Exact Match distance simply counts the number of unequal elements in two permutations, i.e., \smallskip \\ \smallskip
 \centerline{
 $\fdist_{Ham}(\pi,\pi')  = \sum_{i=1}^{m} a_i, ~~\text{where}~~ a_i = \left\{
  \begin{array}{l l}
    0 & \quad \text{if } \pi_i = \pi'_i,\\
    1 & \quad \text{otherwise.}
  \end{array} \right.$}
\item The Euclidean distance is  \smallskip \\ \smallskip
 \centerline{$\fdist_{Euc}(\pi,\pi') =  \sqrt{\sum_{i=1}^{m} (\pi_i-\pi'_i)^2}$ .}
\item The Manhattan distance (A-Distance,\\ cf.~\citep{Sevaux2005,Campos2005}) is \smallskip \\ \smallskip
 \centerline{$\fdist_{Man}(\pi,\pi') =  \sum_{i=1}^{m} |\pi_i-\pi'_i|$ .} 
\item The Chebyshev distance is \smallskip \\ \smallskip
 \centerline{ $\fdist_{Che}(\pi,\pi') =  \underset{1 \leq i \leq m}{\max}(|\pi_i-\pi'_i|)$ .}
\item For permutations, the Lee distance~\citep{Lee1958,Deza1998} is \smallskip \\ \smallskip
  \centerline{$\fdist_{Lee}(\pi,\pi') =  \sum_{i=1}^{m} \min(|\pi_i-\pi'_i|,m-|\pi_i-\pi'_i|)$ .}
\item The non-metric Cosine distance is based on the dot product of two permutations.
It is derived from the cosine similarity~\citep{Singhal2001} of two vectors:
\begin{equation*}
\fdist_{Cos}(\pi,\pi') = 1 - \frac{\pi \cdot \pi'}{||\pi||~||\pi'||}.
\end{equation*}
\item The Lexicographic distance regards the lexicographic ordering of permutations. If the position of a permutation $\pi$
in the lexicographic ordering of all permutations with fixed $m$ is $L(\pi)$, then the Lexicographic distance metric is
\begin{equation*}
\fdist_{Lex}(\pi,\pi') =| L(\pi) - L(\pi')|.
\end{equation*}
\end{itemize}


\section*{Appendix B: Minimal Examples for Indefinite Sets}
To showcase the usefulness of the proposed methods, this section
lists small example data sets and the respective indefinite distance matrices.
Besides the standard permutation distances we also tested:
\begin{itemize}
\item \textbf{Signed permutations, reversal distance:} Permutations where each element has a sign are referred to as signed permutations.
An application example for signed permutations is, e.g., weld path optimization~\citep{Voutchkov2005}. 
The reversal distance counts the number of reversals required to transform one permutation
into another. 
 We used the non-cyclic reversal distance provided in the 
GRAPPA library version 2.0~\citep{Bader2004a}.
\item \textbf{Labeled trees, tree edit distance:} 
Trees in general are widely applied as solution representation, e.g., in Genetic Programming.
In this study, we considered labeled trees. The tree edit distance counts the number node insertions, deletions or relabels.
We used the efficient implementation in the APTED 0.1.1 library~\citep{Pawlik2015,Pawlik2016}. 
The labeled trees will be denoted with the bracket notation: curly brackets indicate the tree structure, letters indicate labels (internal and terminal nodes).
\item \textbf{Strings, Optimal String Alignment distance (OSA):} The OSA is an non-metric edit distance that counts insertions, deletions, substitutions and transpositions of characters. Each substring can be edited no more than once. It is also called the restricted Damerau-Levenshtein distance~\citep{Boytsov2011}. We used the implementation in the \texttt{stringdist} R package~\citep{Loo2014}.
\item \textbf{Strings, Jaro-Winkler distance:}  The Jaro Winkler distance is based on the number of matching characters in two strings as well as the number of transpositions required to bring all matches in the same order.
We used the implementation in the \texttt{stringdist} R package~\citep{Loo2014}.
\end{itemize}

The respective results are listed in Table~\ref{tab:minex}. 
All of the listed distance measures are shown to be non-CNSD. 

%


\begin{table}
\caption{Minimal examples for indefinite distance matrices. The matrix in the table is the actual distance matrix, 
while the eigenvalue refers to the transformed matrix \added{$\hat{D}$} derived from equation~\eqref{eq:augmentedD2}. 
The lower triangular matrix is omitted due to symmetry.
}\label{tab:minex}
\begin{tabular}{lclllll}
\hline
\multicolumn{7}{c}{Permutations, Insert, $n=5$, $m=4$, $\lambda_n \approx$ 0.090} \\ 
\hline
 $i$ & $x_i$ & $d_{i,1}$ & $d_{i,2}$& $d_{i,3}$& $d_{i,4}$&$d_{i,5}$ \\ 
1 &$\{1,2,3,4\}$&0&$1/3$&$1/3$&$2/3$&$1/3$\\
2 &$\{1,3,4,2\}$&&0&$2/3$&$1/3$&$2/3$\\
3 &$\{2,3,4,1\}$&&&0&$1/3$&$2/3$\\
4 &$\{3,4,1,2\}$&&&&0&$1/3$\\
5 &$\{4,1,2,3\}$&&&&&0\\
\hline
\multicolumn{7}{c}{Permutations, Interchange, $n=5$, $m=4$, $\lambda_n \approx$ 0.090} \\
\hline
1 &$\{1,2,3,4\}$&0&$1/3$&$1/3$&$2/3$&$1/3$\\
2 &$\{1,2,4,3\}$&&0&$2/3$&$1/3$&$2/3$\\
3 &$\{1,3,2,4\}$&&&0&$1/3$&$2/3$\\
4 &$\{1,3,4,2\}$&&&&0&$1/3$\\
5 &$\{1,4,3,2\}$&&&&&0\\
\hline
\multicolumn{7}{c}{Permutations, Levenshtein, $n=5$, $m=4$, $\lambda_n \approx$  0.135} \\ 
\hline
1 &$\{1,2,4,3\}$&0&$1$&$1/2$&$1/2$&$1$\\
2 &$\{2,3,1,4\}$&&0&$1/2$&$1/2$&$1$\\
3 &$\{2,4,3,1\}$&&&0&$1$&$1/2$\\
4 &$\{3,1,2,4\}$&&&&0&$1/2$\\
5 &$\{3,4,2,1\}$&&&&&0\\
\hline
\multicolumn{7}{c}{Permutations, LCStr, $n=5$, $m=4$, $\lambda_n \approx$  0.023} \\ 
\hline
1 &$\{1,3,2,4\}$&0&$2/3$&$1/3$&$1/3$&$2/3$\\
2 &$\{2,4,1,3\}$&&0&$1/3$&$1/3$&$2/3$\\
3 &$\{3,2,4,1\}$&&&0&$2/3$&$1$\\
4 &$\{4,1,3,2\}$&&&&0&$2/3$\\
5 &$\{4,2,1,3\}$&&&&&0\\
\hline
\multicolumn{7}{c}{Permutations, Chebyshev, $n=5$, $m=5$, $\lambda_n \approx$ 0.034 } \\ 
\hline
1 &$\{1,5,3,4,2\}$&0&$1/4$&$3/4$&$3/4$&$1$\\
2 &$\{2,5,3,4,1\}$&&0&$1$&$1$&$3/4$\\
3 &$\{4,2,3,1,5\}$&&&0&$2/4$&$1/4$\\
4 &$\{4,3,1,2,5\}$&&&&0&$1/4$\\
5 &$\{5,3,2,1,4\}$&&&&&0\\
\hline
\multicolumn{7}{c}{Sign. Permutations, Reversal, $n=5$, $m=5$, $\lambda_n \approx$ 0.016 } \\ 
\hline
1 &$\{~4,~5,\text{-}1,\text{-}2,\text{-}3\}$&0&$4/6$&$5/6$&$3/6$&$2/6$\\
2 &$\{~2,~1,~3,\text{-}4,\text{-}5\}$&&0&$2/6$&$3/6$&$5/6$\\
3 &$\{\text{-}2,~1,~3,~5,~4\}$&&&0&$5/6$&$3/6$\\
4 &$\{~4,\text{-}2,~3,~1,\text{-}5\}$&&&&0&$2/6$\\
5 &$\{~4,\text{-}2,~1,\text{-}5,\text{-}3\}$&&&&&0\\
\hline
\multicolumn{7}{c}{Labeled Trees, Edit dist., $n=5$, $\lambda_n \approx$ 0.026} \\ 
\hline
1 &$\{b\{c\{b\}\}\}$&0&$2$&$1$&$3$&$1$\\
2 &$\{b\}$&&0&$1$&$3$&$1$\\
3 &$\{b\{c\}\}$&&&0&$2$&$2$\\
4 &$\{a\{c\}\{a\}\}$&&&&0&$3$\\
5 &$\{c\{b\}\}$&&&&&0\\
\hline
\multicolumn{7}{c}{Strings, Optimal String Alignment, $n=5$, $\lambda_n \approx$ 0.102 } \\ 
\hline
1 &abc&0&$1$&$2$&$3$&$1$\\
2 &acc&&0&$3$&$2$&$2$\\
3 &cba&&&0&$1$&$2$\\
4 &caa&&&&0&$2$\\
5 &bac&&&&&0\\
\hline
\multicolumn{7}{c}{Strings, Jaro-Winkler, $n=4$, $\lambda_n \approx$ 0.046 } \\ 
\hline
1 &bbbb&0&$1$&$1/6$&$3/6$&\\
2 &aaaa&&0&$3/6$&$1/6$&\\
3 &bbba&&&0&$3/6$&\\
4 &aaab&&&&0&\\
\hline
\end{tabular}
\end{table}

\end{document}